\documentclass{article}

\usepackage[final]{infer2control_2018}

\usepackage[utf8]{inputenc} 
\usepackage[T1]{fontenc}    
\usepackage{hyperref}       
\usepackage{url}            
\usepackage{booktabs}       
\usepackage{amsfonts}       
\usepackage{nicefrac}       
\usepackage{microtype}      
\usepackage{times}
\usepackage[small]{caption}
\usepackage{amsmath} 
\usepackage{amssymb} 
\usepackage{caption}
\usepackage{breqn}
\usepackage[mathscr]{euscript}
\usepackage{microtype}
\usepackage{graphicx}
\usepackage{subfigure}
\usepackage{booktabs} 
\usepackage{color}

\usepackage{authblk}

\title{Learning with Stochastic Guidance for Navigation}

\author[1]{Linhai Xie}
\author[1]{Yishu Miao}
\author[2]{Sen Wang}
\author[1, 3]{Phil Blunsom}
\author[1]{Zhihua Wang}
\author[1]{Changhao Chen}
\author[1]{Andrew Markham}
\author[1]{Niki Trigoni}
\affil[1]{University of Oxford}
\affil[2]{Heriot-Watt University}
\affil[3]{DeepMind}

\begin{document}

\maketitle

\begin{abstract}
Due to the sparse rewards and high degree of environment variation, reinforcement learning approaches such as Deep Deterministic Policy Gradient (DDPG) are plagued by issues of high variance when applied in complex real world environments. We present a new framework for overcoming these issues by incorporating a stochastic switch, allowing an agent to choose between high and low variance policies.
The stochastic switch can be jointly trained with the original DDPG in the same framework.
In this paper, we demonstrate the power of the framework in a navigation task, where the robot can dynamically choose to learn through exploration, or to use the output of a heuristic controller as guidance.
Instead of starting from completely random moves, the navigation capability of a robot can be quickly bootstrapped by several simple independent controllers.
The experimental results show that with the aid of stochastic guidance we are able to effectively and efficiently train DDPG navigation policies and achieve significantly better performance than state-of-the-art baselines models.
\end{abstract}

\section{Introduction}

Deep Reinforcement Learning (DRL) has been shown to be extremely effective in mastering complex simulations and artificial tasks, e.g. playing Atari games [13] and Go [17]. However, DRL's poor sample complexity has limited its application to real world tasks, such as navigating a robot to a target position without crashing.

Deep Deterministic Policy Gradient (DDPG) [10] is an actor-critic algorithm that is suitable for such continuous control tasks in principle, but in practise the cost of exploration in complex navigation environments can prove prohibitive.

Since an agent must stochastically explore a long sequence of states during each training episode, high variance becomes the main bottleneck that hinders DDPG from learning effective DRL models. In order to mitigate this issue, conventional architectures generally require a huge number of learning samples,  resulting in high computational and environmental costs in practice. 
In this paper, we propose a new framework that allows an agent to stochastically switch between high variance controllers (e.g. DDPG), and low variance controllers (e.g. simple deterministic controllers), effectively allowing the DDPG component to be quickly bootstrapped instead of starting from completely random moves.

Intuitively, learning is usually easier to be carried out under the guidance from other heuristics.
The independent controllers here act as the guidance that are introduced for learning better DDPG policies.
In our case, the agent still maintains an independent DDPG module that learns navigation by exploring the environment, but it is able to dynamically switch between learning from exploration or learning from the heuristic controllers.
Here, the switching mechanism is constructed as a stochastic function updated by REINFORCE learning signal [23] to maximise total reward.
Meanwhile, the DDPG component is learned by employing the action selected by the stochastic switch, rather than directly using the output action generated by its policy network.
Therefore, the switching mechanism helps DDPG avoid trivial explorations during the early training process, and learns to balance between exploration and heuristic guidance. 
More interestingly, the DDPG component can be tested in isolation from other controllers, in which case the switch is turned off and the navigation is carried out solely by the DDPG component. 
Similar to the idea of imitation learning [3], the DDPG component is able to learn from the demonstrations given by the guidance (PID and OA (obstacle avoidance) components in our case) and instantly generalise to new situations (which PID and OA could not handle). 
Here, the guidance can be considered as a positive bias for reducing the variance of gradient estimators, and the model is able to remove this bias after benefiting from it.

For quantitative evaluation, we firstly compare our model with stochastic switch to the vanilla DDPG baseline and deterministic benchmarks for demonstrating the benefits brought by independent controllers.
Then the influence of using different independent controllers is investigated, which shows that the framework has strong generalisation ability and it is able to accumulate the benefits from different simple controllers.
In addition, we propose two variants of the switch mechanism including a uniformly random switch and an argmax switch for comparison.
Finally, we show that the models can abandon the extra controllers when their usage rate declines below a threshold, and it is able to continue the self-learning by only using the DDPG component.
For qualitative evaluation, we test our model in a real world scenario. Without further modification, the model trained in simulation is able to be directly transferred to carry out navigation tasks.

In summary, we propose a new framework that leverages the heuristic knowledge provided by independent controllers to bootstrap deep reinforcement learning for robot navigation.
Our experiments demonstrate that by incorporating stochastic guidance, we are able to effectively and efficiently train the DDPG navigation policies and achieve significantly better performance than state-of-the-art baseline models.
As a simple, robust and easy-to-use framework, it can be a generic method applied to improve many other deep reinforcement learning algorithms and applications.

\begin{figure}[!t]
  \vspace{-0.2cm}
  \centering
  \includegraphics[width=0.9\linewidth]{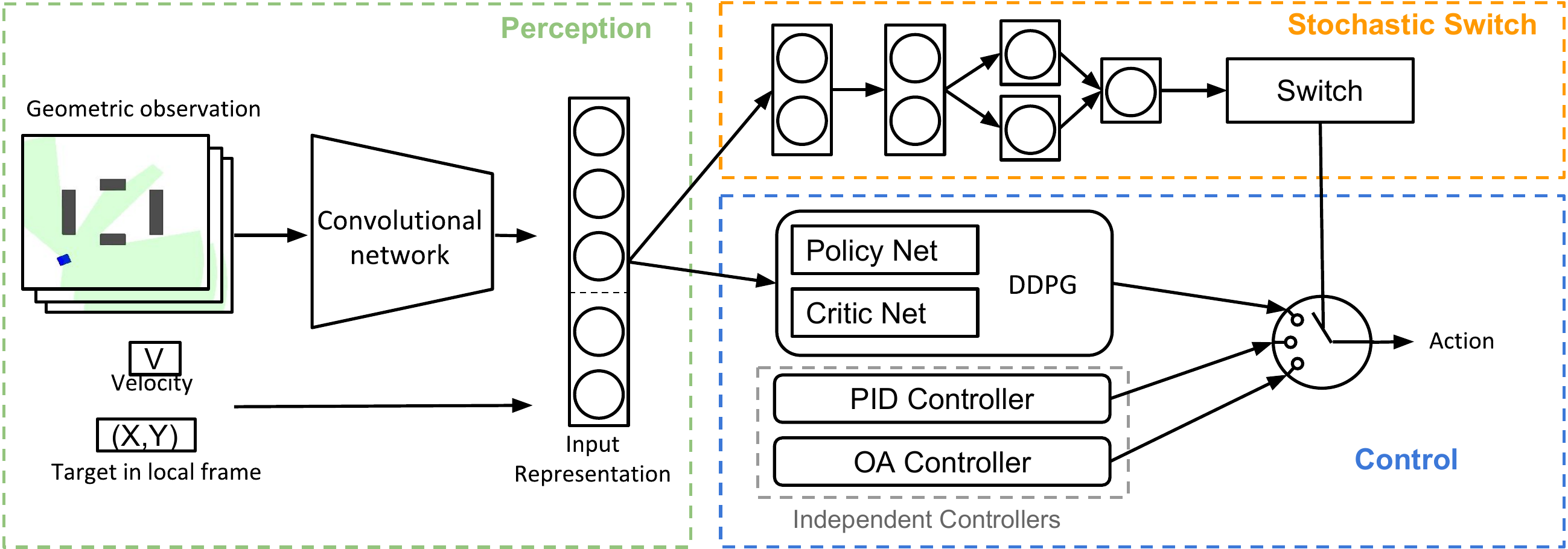}
  \caption{The architecture of proposed framework. It consists of three sections including perception, controllers and stochastic switch. 
  The perception part processes an observation and generates a corresponding input representation. 
  The controllers part contains two independent controllers (Proportional-integral–derivative (PID) controller and Obstacle Avoidance (OA) controller) and a DDPG controller.
  The stochastic switch determines which one of the actions proposed by the controllers to be eventually carried out for navigation.}
  \label{fig:Network}
  \vspace{-0.1cm}
\end{figure}

\section{Model}

Robot navigation task can be defined as a partially observable Markov decision process [18] problem, which can be solved by DRL. With observations of world state, a robot needs to decide its action, i.e. a control policy, maximising an accumulative future reward. Since it is ideal for a robot to reach the goal without any crash, the reward function $r_t$ at time $t$ is defined as:
\begin{equation}
    r_t = 
\begin{cases}
    R_{crash}, & \text{if robot crashes}\\
    R_{reach}, & \text{if robot reaches the goal}\\
    (d_{t-1} - d_t)\cos{(\omega_t)} - C, & \text{otherwise}
\end{cases}
\label{eq:reward function}
\end{equation}

where $R_{crash}$ is a large penalty for collision, $R_{reach}$ is a positive reward for reaching the goal, $d_{t-1}$ and $d_t$ denote the distances between the robot and the goal at two consecutive time steps $t-1$ and $t$, $\omega_t$ represents the rotational speed of the robot at time $t$, and $C$ is a constant time penalty which encourages the robot to approach the goal quickly.

The proposed model consists of three parts: perception, control and stochastic switch, as shown in Fig.\ref{fig:Network}. 
At each time step, the perception part processes an observation and generates a corresponding input representation. Then different controllers can propose candidate actions based on the input representation.
Finally, the stochastic switch determines which one of the actions to be carried out.





\subsection{Perception}

At each time step $t$, the robot observes the state of the world $x_t$, which includes a stack of current and historical geometric observations, its linear and angular velocities and a destination. The geometric observations, which give distances to surrounding objects (depth images in this work), are processed by a convolutional neural network to produce a compressed input representation. It is then concatenated with the velocities and destination for controllers and stochastic switch.

\subsection{Control}

\paragraph{Action} With the observation $x_t$ at time $t$, the robot takes an action $a_t = (a^v_t, a^{\omega}_t) \in \mathscr{A}$, where $a^v_t$ and $a^{\omega}_t$ respectively denotes the expected linear and rotational velocity at time $t$, to navigate. Then it obtains a reward $r_t$ given by the environment for assessing the chosen action and transits to next observation $x_{t+1}$. The goal of our model is to reach a maximum discounted accumulative future reward $R_t = \sum_{t'=t}^{T}\gamma^{t'-t}r_t$, where $\gamma$ is the discount factor and $r_t$ is the reward function in Eq. \ref{eq:reward function}. 
In this work, the actions can be determined by the independent controllers and DDPG, establishing a set of candidate actions for the stochastic switch to choose.

\paragraph{Independent Controllers} Two independent controllers are introduced to facilitate the learning of DDPG's policies, especially providing reasonable actions in the beginning. One is a proportional-integral-derivative (PID) controller with proportional term [2], which derives action from the relative position of the destination $[x_{local}, y_{local}]$ in robot coordinate frame as:
\begin{equation}
a_t = K_p \cdot [x_{local}, y_{local}]^T,
\end{equation}
where $K_p$ is the coefficient for the proportional term. PID controller is one of the most widely used and successful control mechanisms. However, without considering geometric observation, it does not have an obstacle avoidance capability.

The other one is a simple obstacle avoidance (OA) algorithm which can drive the robot without collision. It uses geometric observation to detect and avoid nearby obstacles by controlling the heading direction (rotational speed) of the robot:
\begin{equation}
    |a^{\omega}| = 
\begin{cases}
    a^{\omega}_{max}\cdot\frac{|d_{o} - \beta|}{\beta}, & d_{o} < \beta\\
    0, & \text{otherwise.}\\
\end{cases}
\label{eq:OA controller}
\end{equation}
where $d_{o}$ is the distance to the closest obstacle, $a^{\omega}_{max}$ represents the largest rotational speed and $\beta$ indicates a pre-defined minimum safety distance. 
In the case where the distance between the robot and an object is less than the safety distance, i.e., $d_{o} < \beta$, the robot will rotate to avoid collision.
These two controllers complement one another to provide candidate actions for stochastic switch.
Note that the OA only produces $a^{\omega}$, while the selected $a^v$ is provided by the DDPG controller.

\paragraph{DDPG} The main controller of this framework is DDPG, which is an actor-critic approach in DRL [10] that simultaneously learns the policy and the action-state value (Q-value) to assess the learnt policy. Although the policy network and the critic network of the DDPG share the same input representation from the perception, the policy network predicts the action, while the critic network estimates the Q-value for current state-action pair.
In the learning mechanism of critic network, given the policy $\pi$ which maps states to actions $a_t$, the expected return is
$Q^\pi(x_t,a_t) = \mathbb{E}[R_t|x_t,a_t,\pi],$
which can be calculated with the Bellman equation [22]:
\begin{equation*}
	Q^\pi(x_t,a_t) = \mathbb{E}[r_t+\gamma\mathbb{E}[Q^\pi(x_{t+1},a_{t+1})|x_t,a_t,\pi].
\end{equation*}
If the policy is deterministic, we can define $a_t=\mu(x_t)$ and the inner expectation can be avoided.
Since the outer expectation is independent on policy $\mu$, it becomes an off-policy learning.
Then, the objective is to minimise the temporal difference (TD) error: 
\begin{equation}
  \begin{aligned}
    L(\theta^Q)&=\mathbb{E}_{x_t,a_t,r_t,x_{t+1}}[(y - Q(x_t,a_t;\theta^Q))^2], \\
    y &= r_t+\gamma Q(x_{t+1},a_{t+1};\theta^Q)
  \end{aligned}
\end{equation}
where $\theta^Q$ is a parameter of the critic network. To update the critic network by temporal difference learning [21], all learning samples stored in the replay buffer are formulated as $(x_t, a_t, r_t, x_{t+1})$.

The policy network is parameterised by $\theta ^\mu$. During training, the gradients are estimated by applying chain rules to the objective function (expected reward) $J(\theta^{\mu})$ w.r.t the parameters $\theta^\mu$.
Generally, in DDPG, the parameters are updated by the gradients computed based on the actions produced by the policy network.
However, in our case, we introduce a stochastic switch for choosing the final action from a set of actions proposed by all controllers.
Hence, the networks are actually updated by the action decided by the switch network, instead of the $a_t$ produced by the policy network of DDPG.

\subsection{Stochastic Switch}

The PID controller, OA algorithm and DDPG are three independent sources that produce candidate actions for the switch network to (optimally) select. The switch network is a stochastic deep neural network which consists of a parameterisation network and a multinomial distribution. 
Conventionally, a softmax layer can be employed to provide the parameter $\theta$ for the multinomial distribution.
Here, instead, we apply \textbf{stick-breaking construction} [16, 8, 11], which is alternative to softmax.

The intuition is to introduce a bias that encourages more usage of the deep reinforcement learning algorithm, such as DDPG in our case.
Since our framework is designed to train a robust DDPG component that benefits from the stochastic guidance, we expect it to be used more often than others in this framework so that we are able to get rid of the simple independent controllers after a certain period of training.

It basically transforms the modelling
of multinomial probability parameters into the modelling of the logits of binomial probability parameters. In our case ($K=3$ controllers), given the binomial logits $\eta = \text{sigmoid}(\alpha)$, $\xi$ can be generated by two breaks: $\xi_1=\eta_1$, $\xi_2=\eta_2(1-\eta_1)$ and $\xi_3=(1-\eta_2)(1-\eta_1)$.
Here $\alpha = f_s(x_t|\theta^s)$ is the unscaled logit from a deep neural network $f_s(\cdot|\theta^s)$ given the input representation $x_t$ and $\theta^s$ is the parameter of the switch network.
Generally, the stick breaking function $f_{\text{SB}}$ can be generalised to more breaks $\xi_k = \eta_k \prod\nolimits_{i=1}^{k-1}(1-\eta_i)$ ($1<k<K$).
Conditioned on the current observation $x_t$, we are able to construct the stochastic switch policy $s_t \sim \pi^s(s_t|x_t;\theta^s)$ as:
\begin{align}\label{eq: stick-breaking}
 \xi &= f_{\text{SB}}(\text{sigmoid}(f_s(x_t|\theta^s))) \\
 s_t &\sim  \text{multinomial}(\xi).
\end{align}
At each time step $t$, the stochastic switch samples a decision $s_t$ and $\theta_1, \theta_2, \theta_3$ corresponds to DDPG, PID and OA.
Then, according to the decision $s_t$, the critic network of DDPG takes the final action $a_{s_t} \in \{a_{\text{DDPG}}, a_{\text{PID}}, a_{\text{OA}}\}$ as input and updates the networks accordingly.
Meanwhile, the stochastic switch is updated by REINFORCE learning signal so that the switch network is able to dynamically choose to learn through exploration (DDPG), or it can choose to use the output of a heuristic controller (PID or OA) as guidance by observing the environment.

\paragraph{REINFORCE Algorithm}
Since the gradients cannot be directly back-propagated through the discrete samples, we employ REINFORCE algorithm [23] to construct the gradient estimator for the switch network, where the goal is to maximise the total reward $R$ under the switching policy $\pi^s(s_t|x_t;\theta^s)$. Thus the objective function is:
\begin{equation}
 \mathcal{H} = \mathbb{E}_{p(S; \theta^s)}[R] = \mathbb{E}_{p(S; \theta^s)}\Big[ \sum\nolimits_{t=1}^T \gamma^{t-1} r_t \Big] ,
\end{equation}
where $S$ is a sequence of decisions $ s_1, s_2, ..., s_T$ in an episode, $s_t \sim \pi^s(s_t|x_t;\theta^s)$ is the decision sample at each time step $t$, and $p(S; \theta^s) = \prod_{t=1}^T \pi^s(s_t|x_t;\theta^s) $ is the probability of generating the current decision sequence.
Hence, the gradients can be estimated as follows:
\begin{dmath}\label{eq:reinforce}
    \dfrac{\partial \mathcal{H} }{\partial \theta^s}=\mathbb{E}_{p(S; \theta^s)}\Big[\dfrac{\partial}{\partial \theta^s}\log p(S;\theta^s) R\Big] \approx 
    \dfrac{1}{N}\sum_{n=1}^N\sum_{t=1}^{T_n} \dfrac{\partial}{\partial \theta^s}\log \pi^s(s_t^n|x_t^n;\theta^s)R^n
\end{dmath}
where $N$ is the number of sampled episodes, $T_n$ is the length of the episode $n$, and $R^n$ is the total reward of the episode. 
It indicates a Monte Carlo based unbiased gradient estimation for updating the switch network. 

The introduction of stochastic switch can be considered as an inductive bias for learning to navigate with better action samples.
Updated by REINFORCE, the stochastic switch is able to sense the environment, avoid the trivial explorations and select better actions for learning DDPG policies.
In addition, as the independent controllers are incorporated via the stochastic switch, the negative influence of the introduced biases from the heuristics is limited.
More interestingly, the independent controllers can be turned off in the late training process (the stochastic switch always chooses the output of DDPG as final action) so that the learning of DDPG could further rely on exploration after being bootstrapped by the independent controllers. 

\paragraph{Variance Reduction}
Since the REINFORCE gradient estimator also suffers from the high variance issue, 
we introduce two control variates [12] for alleviating the problem: a centred learning signal $b^c$ (moving average) and an input dependent
control variate $b(x)$ respectively.
Here, we simply build an MLP (multilayer perceptron) to implement the $b(x)$ conditioned on input $x$. 
During training, the two control variates are learned by minimising the expectation:
$\mathbb{E}_{p(S; \theta^s)}[R-b^c-b(x))]^2$, and the gradients are derived as,
\begin{equation}\label{eq:baseline}
\dfrac{\partial \mathcal{H} }{\partial \theta^s} = \dfrac{1}{N}\sum_{i=1}^N\sum_{t=1}^{T_n} \dfrac{\partial}{\partial \theta^s}\log \pi^s(s_t^n|x_t^n;\theta^s)(R^n-b^c-b(x_t)) \nonumber
\end{equation}

\section{Experiments}
\vspace{-0.2cm}
\begin{figure}[!tb]
\vspace{-0.2cm}
\centering
\subfigure[\textit{ROS Stage}]{\label{fig:stage}\includegraphics[width=27mm]{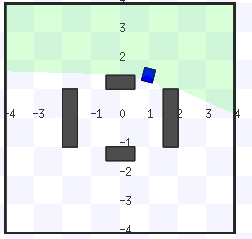}}
\subfigure[\textit{ROS Gazebo}]{\label{fig:gazebo}\includegraphics[width=30mm]{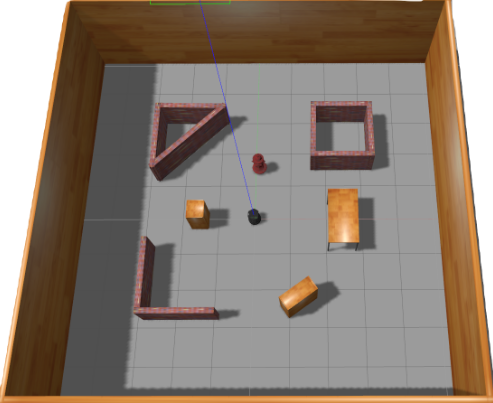}}
\caption{
\textbf{(a)} The 4 grey rectangles are obstacles and the blue square represents the robot. A sparse laser is mounted on the robot and its detecting area is illustrated as the green area. 
\textbf{(b)} It shows a more complex environment simulated by \textit{ROS Gazebo}.
Turtlebot 2 (a platform for ground robots) is employed as the mobile platform equipped with a depth camera.
}
\vspace{-0.2cm}
\label{fig:simulated_world}
\end{figure}
\subsection{Training Environments and Settings}

The proposed framework is trained in two different simulators. 
The first one is a light-weight simulator, \textit{ROS Stage}\footnote{http://wiki.ros.org/stage} (Fig.\ref{fig:stage}), in which a large amount of repetitive experiments are conducted for showing the learning curve, demonstrating the improvements brought by stochastic guidance, and comparing to other baseline models. 
In this simulator, we mount the mobile robot with a laser scanner to provide the geometric information of surroundings.
Hence, the convolutional neural network (in Fig. \ref{fig:Network}) is not being used in this case, and the laser scans are directly concatenated with other observation as input representation.
By accelerating the simulation time, we obtain the \textbf{quantitative evaluation} through a lot of repetitive experiments in \textit{ROS Stage}.

The other one, \textit{ROS Gazebo}\footnote{http://wiki.ros.org/gazebo_ros_pkgs} (Fig.\ref{fig:gazebo}), contains a physical engine and can accurately simulate the dynamics of the mobile robot. 
Thus the model trained in \textit{ROS Gazebo} is directly applied to real world scenario to \textbf{qualitatively evaluate} the navigation performance, but it has a larger computational overhead compared to \textit{ROS Stage}. 
Here, depth images are utilised to observe surroundings, therefore a 3-layer convolutional network (the filters are [4,4,3,8], [4,4,8,16] and [4,4,16,32] respectively) is constructed to provide input representations based on depth images.

In each training episode the robot starts at the origin point with a random heading direction and the destination is randomised within the area beyond obstacles. 
When the robot collides with an obstacle or reaches the destination, the current episode terminates. 
The action control frequency is 5Hz 
and the switching frequency is 1Hz
.
For all the experiments carried out in \textit{ROS Stage}, the training process lasts for 100k steps and is repeated for 5 times. 
The averaged learning curves as well as the variance\footnote{Note that the variance mentioned here is the variance of the smoothed learning curves.} are illustrated for demonstrating the performance.

As for the hyper-parameters, 
the hidden layers of critic network and actor network contain 100 ReLU units in each layer, while the output layer of actor network applies tanh and sigmoid respectively for rotational and linear velocity. 
When updating DDPG parameters, 32 learning samples are randomly sampled from a rank based prioritised experience replay [15] as a training batch, and the learning rate for the actor network, the critic network and the stochastic switch are $10^{-4}$, $10^{-3}$ and $10^{-3}$ respectively, and the rest follows [10].

\subsection{Navigation in Simulated Environment}

\paragraph{Reinforcement Learning with Stochastic Guidance}
\begin{figure}[!tb]
    \centering     
    \subfigure[Comparison with baselines and oracle]{\label{fig:compare_approaches}\includegraphics[width=55mm]{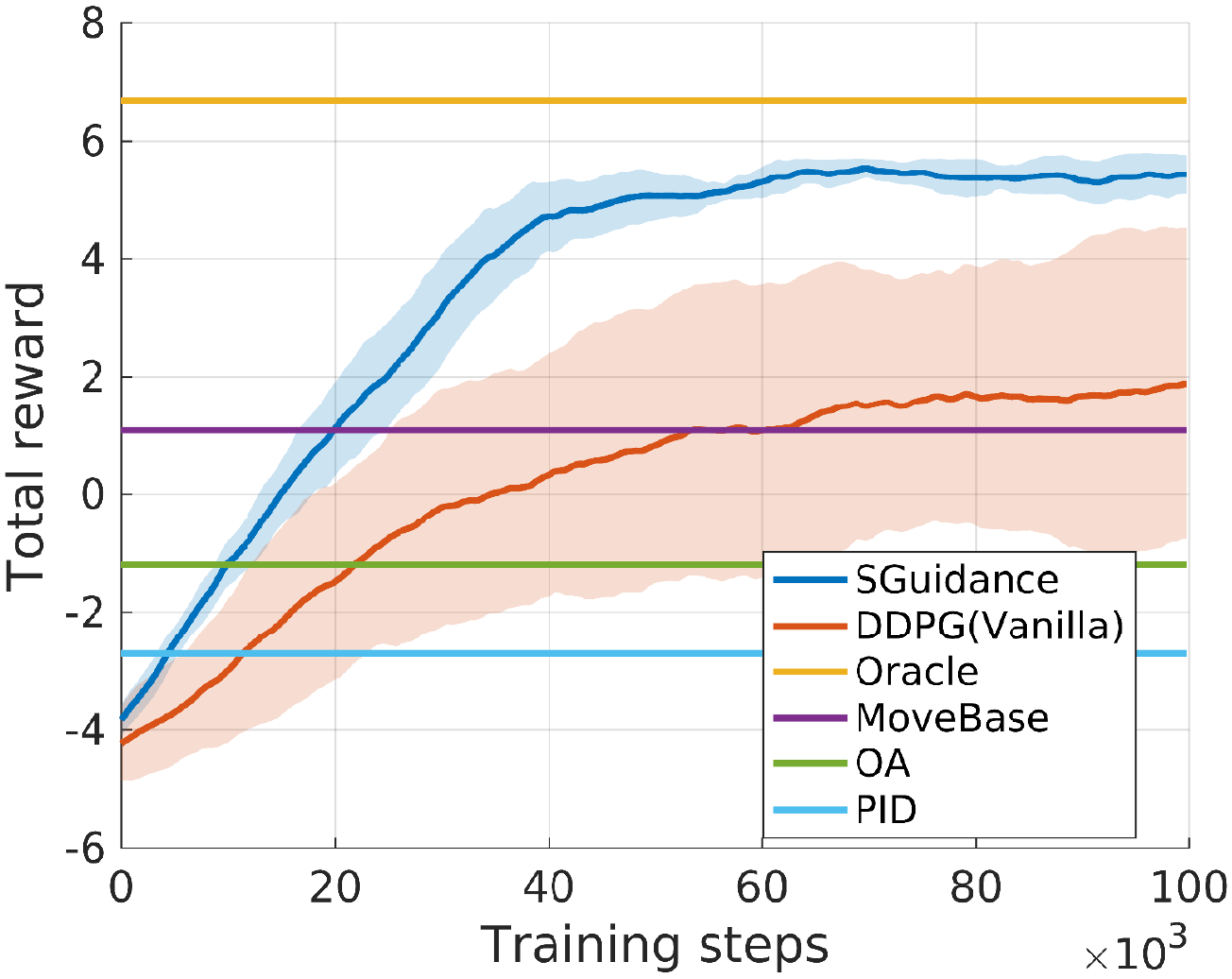}}
    \hspace{1cm}
    \subfigure[Using different independent controllers]{\label{fig:compare_configurations}\includegraphics[width=55mm]{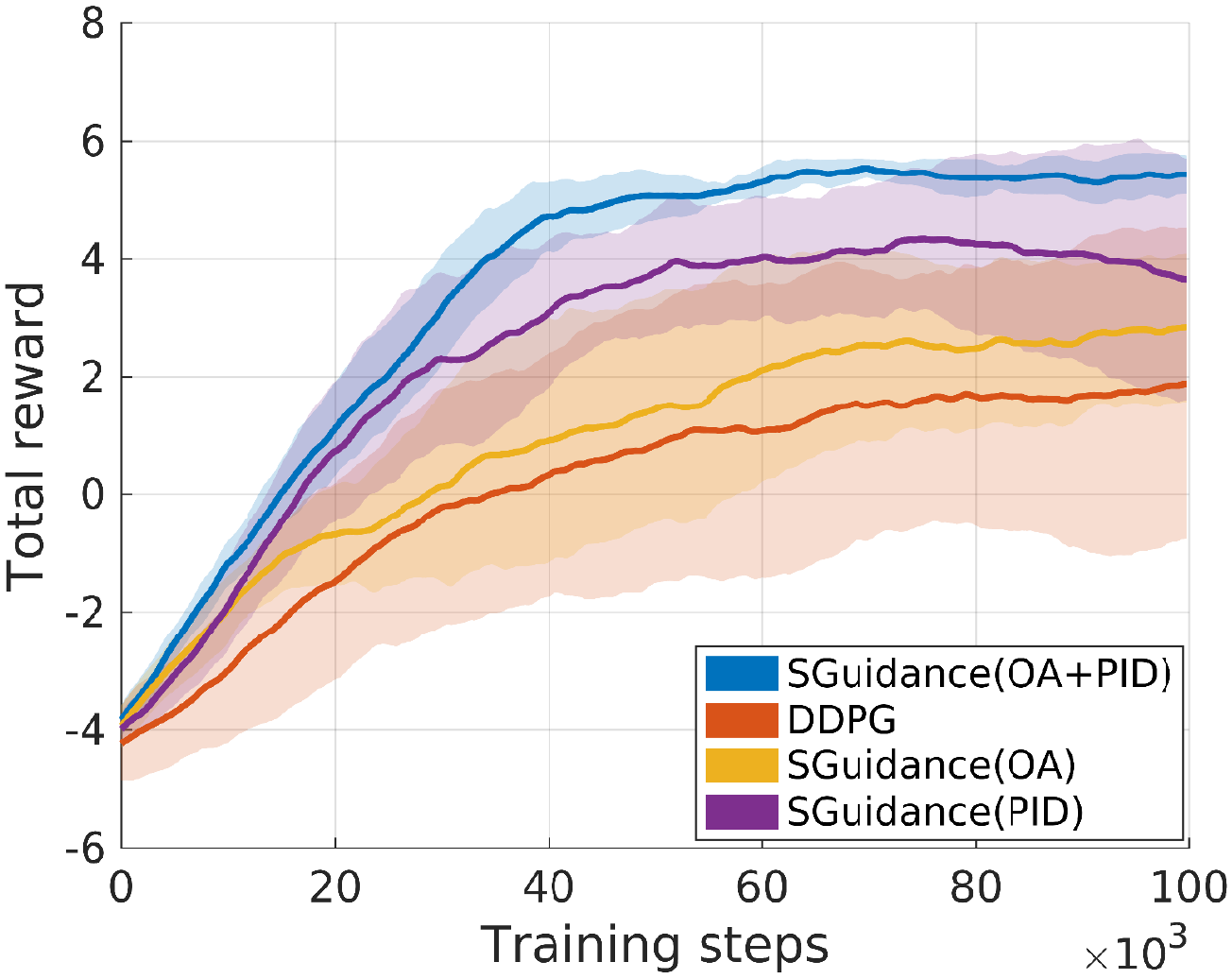}}
    \caption{
    \textbf{(a)} The total reward achieved by our models and other baseline models as comparison. 
    The curve represents the average value of 5 repetitive training procedures and the transparent area indicates the variance of the results.
    \textbf{(b)} The smoothed total reward obtained by incorporating different heuristic controllers with DDPG.
    \textbf{SGuidance} utilises both PID and OA controllers while \textbf{DDPG+OA} and \textbf{DDPG+PID} only adopt one of them respectively. 
    \textbf{DDPG(Vanilla)} is the baseline DDPG without heuristic guidance.}
\end{figure}

Fig.\ref{fig:compare_approaches} compares the models for demonstrating the benefits brought by learning with stochastic switch.
\textbf{SGuidance} is our model with stochastic switch that dynamically choose the action from the candidates proposed by the controllers of DDPG, PID and OA. 
As shown in the figure, \textbf{SGuidance} achieves significantly better performance than the \textbf{DDPG} baseline. 
Meanwhile, \textbf{DDPG} suffers from the high variance issue according to the wide transparent area around the learning curve, while \textbf{SGuidance} is much more stable. 
This is due to the high complexity of the environment that leads to the highly variant learning samples provided by DDPG, which might lead to trivial explorations.
In addition, the stochastic gradient estimator of DDPG applies biased approximation which makes it difficult to guarantee the convergence and stability.
By contrast, \textbf{SGuidance} is able to benefit from the heuristic simple controllers since the beginning of training procedure instead of starting from completely random moves. 

In this experiment, we also plot the rewards of \textbf{MoveBase}  (without map) and \textbf{Oracle} (\textbf{MoveBase} with map) for comparison.
\textbf{MoveBase} package\footnote{http://wiki.ros.org/move_base} is a widely used motion planner for mobile robot navigation and is implemented in the ROS package named Navigation Stage, which consists of a local planner [5, 4] and a global planner (implemented by Dijkstra or A* algorithm).
The global planner generates an optimal path from the origin to the destination on the global map of the environment, and the local planner dynamically avoids the newly detected obstacles while moving along the optimal path.
Hence, we call the \textbf{MoveBase} with map \textbf{Oracle} in this experiment.
As shown in Fig.\ref{fig:compare_approaches}, 
\textbf{DDPG} is able to obtain comparable performance to \textbf{MoveBase}. 
\textbf{SGuidance}, however, significantly surpasses the deterministic \textbf{MoveBase} model.
Even without the access to the global map, \textbf{SGuidance} has shown its strong ability to navigate in the environment by just using the geometric information.
In addition, we plot the performance of two simple heuristic controllers (OA and PID) for reference.
Basically, the simple deterministic controllers can not be applied independently for carrying out navigation task (the accumulative rewards are both under 0). 
However, when incorporated with DDPG via stochastic switch, it contributes notably for alleviating the high variance issue during the learning of DDPG.


\paragraph{Using Different Independent Controllers}

This experiment shows the investigation where different independent controllers are incorporated with DDPG via stochastic switch.
As illustrated in Fig.\ref{fig:compare_configurations}, \textbf{SGuidance (PID + OA)} achieves the best performance when compared to the DDPG with only PID or OA and the DDPG without any independent controllers.
Moreover, the contribution of the stochastic switch is greatly enhanced by adding more controllers, which yields more stable learning curves and better navigation performance.
Interestingly, PID controller brings more benefits than OA controller in this context, and their benefits could be accumulated with the help of the stochastic switch.

\paragraph{Using Different Switching Mechanism}

\begin{figure}[!tb]
    \vspace{-0.2cm}
    \centering     
    \subfigure[Using different switching mechanism]{\label{fig:deterministic}\includegraphics[width=54mm]{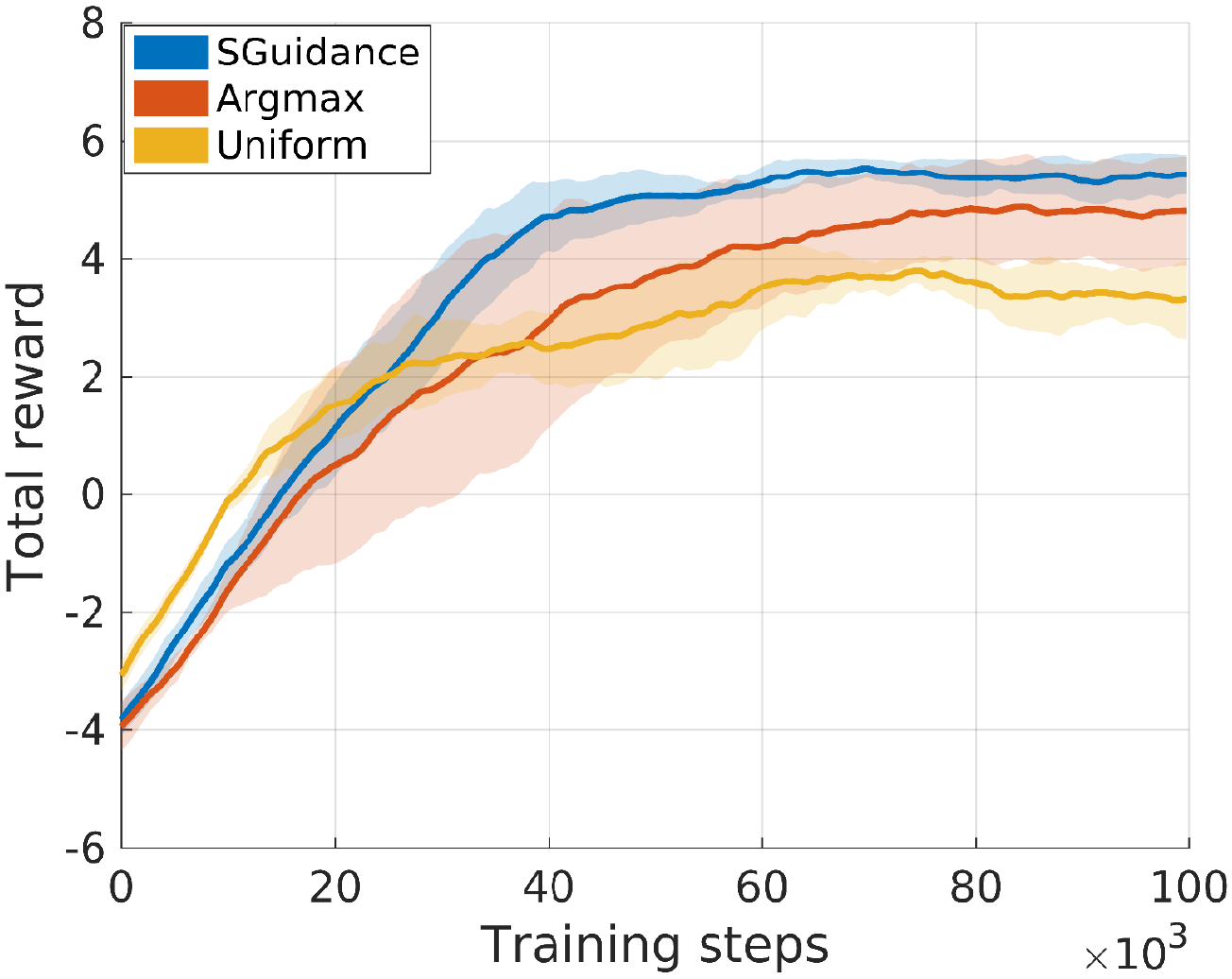}}
    \hspace{1cm}
    \subfigure[Using different switching function]{\label{fig:sup1}\includegraphics[width=57mm]{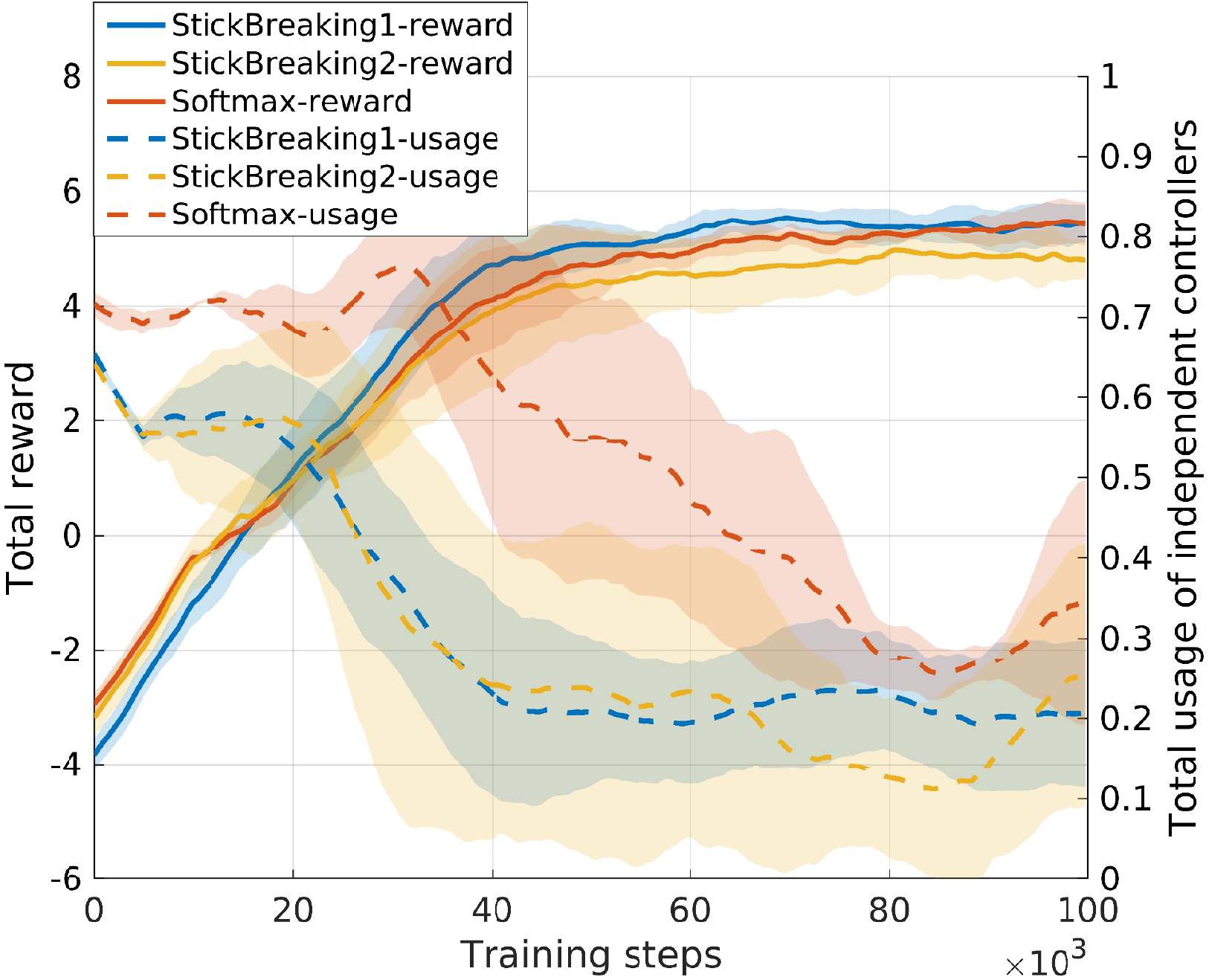}}
    \caption{
    \textbf{(a)} The accumulated reward obtained by learning with different switching mechanism. \textbf{SGuidance} represents our default stochastic switch settings. \textbf{Argmax} and \textbf{Uniform} are the proposed argmax switch and uniformly distributed switch respectively. 
    \textbf{(b)} Discussion on different stochastic switch function. The left y-axis shows the total reward of all the methods, and the right y-axis shows the total usage of the independent controllers.
    }
\vspace{-0.2cm}    
\end{figure}


Fig.\ref{fig:deterministic} compares the stochastic switch to other variants of switching mechanism, an uniformly random switch and an argmax switch.
The uniform switch assigns uniformed fixed probability to DDPG, PID and OA controllers, while the argmax switch applies biased argmax output instead of stochastically drawing samples from the stochastic switch network.
As illustrated in Fig.\ref{fig:deterministic}, \textbf{SGuidance} has the best performance. 
\textbf{Uniform} switch is not as good as \textbf{SGuidance}, but it still contributes remarkably to the navigation performance.
The curve of \textbf{Argmax} lies in between the other two mechanisms, but has much bigger variance on the total reward. 
This is because \textbf{Argmax} is a biased sampler compared to the others, and the introduced bias in turn damages its final performance since there are less explorations after a certain period of training.



\paragraph{Construction of Stochastic Switch Function}

In Fig. \ref{fig:sup1}, the \textbf{StickBreaking1} (DDPG, PID, OA) represents the function we applied in the paper and the \textbf{StickBreaking2} (DDPG, OA, PID) used an alternative order of the independent controllers.
More specifically, according to Eq. \ref{eq: stick-breaking}, \textbf{StickBreaking1} and \textbf{StickBreaking2} both set $\eta_1$ with DDPG controller and give different order with PID and OA controller where $\eta_2$ is assinged with the PID controller in \textbf{StickBreaking1} but with OA controller in \textbf{StickBreaking2}.
As shown in the figure, \textbf{Softmax} is able to achieve almost adequate performance compared to \textbf{StickBreaking1} in terms of the total reward.
However, according to the total usage of independent controllers, the DDPG component is being less used in \textbf{Softmax} than \textbf{StickBreaking1} and \textbf{StickBreaking2}.
Although the two stick breaking functions have similar total usage of independent controllers, \textbf{StickBreaking2} performs slightly worse than \textbf{StickBreaking1}, which shows that the order of independent controllers has a small effect on the performance.
Hence, the softmax function is a safe choice to construct the stochastic switch function. However, the prior knowledge about the performance of simple controllers could be used to benefit the learning in stick breaking construction. For instance, Fig. \ref{fig:compare_configurations} shows PID brings more benefit than OA when incorporating with stochastic switch, and Fig. \ref{fig:sup1} also shows the \textbf{StickBreaking1} performs slightly better than \textbf{StickBreaking2}.

\begin{figure}[!tb]
    \vspace{-0.2cm}
    \centering     
    \subfigure[Variance reduction]{\label{fig:variance reduction}\includegraphics[width=54mm]{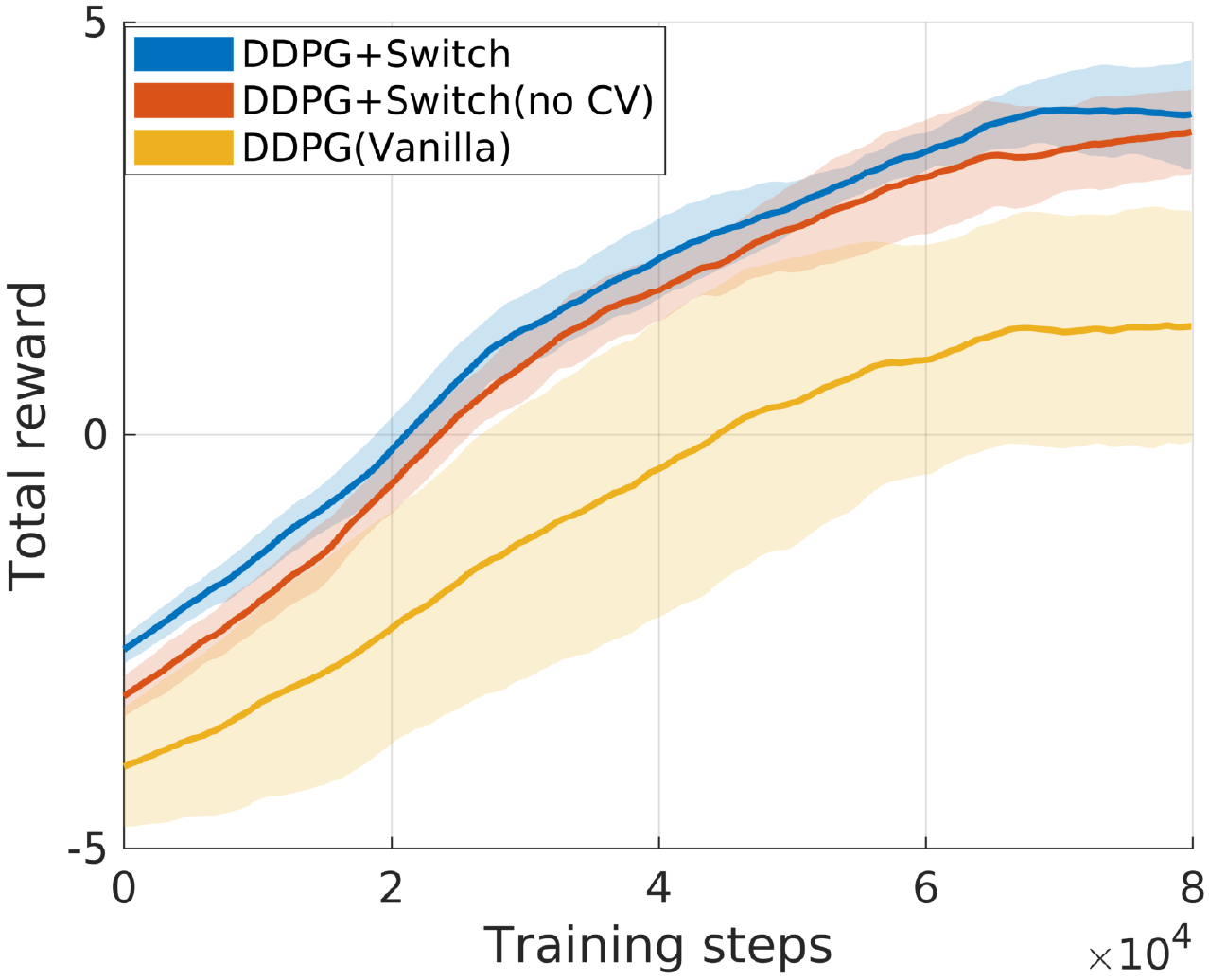}}
    \hspace{1cm}
    \subfigure[Turning off simple controllers]{\label{fig:switchoff}\includegraphics[width=57mm]{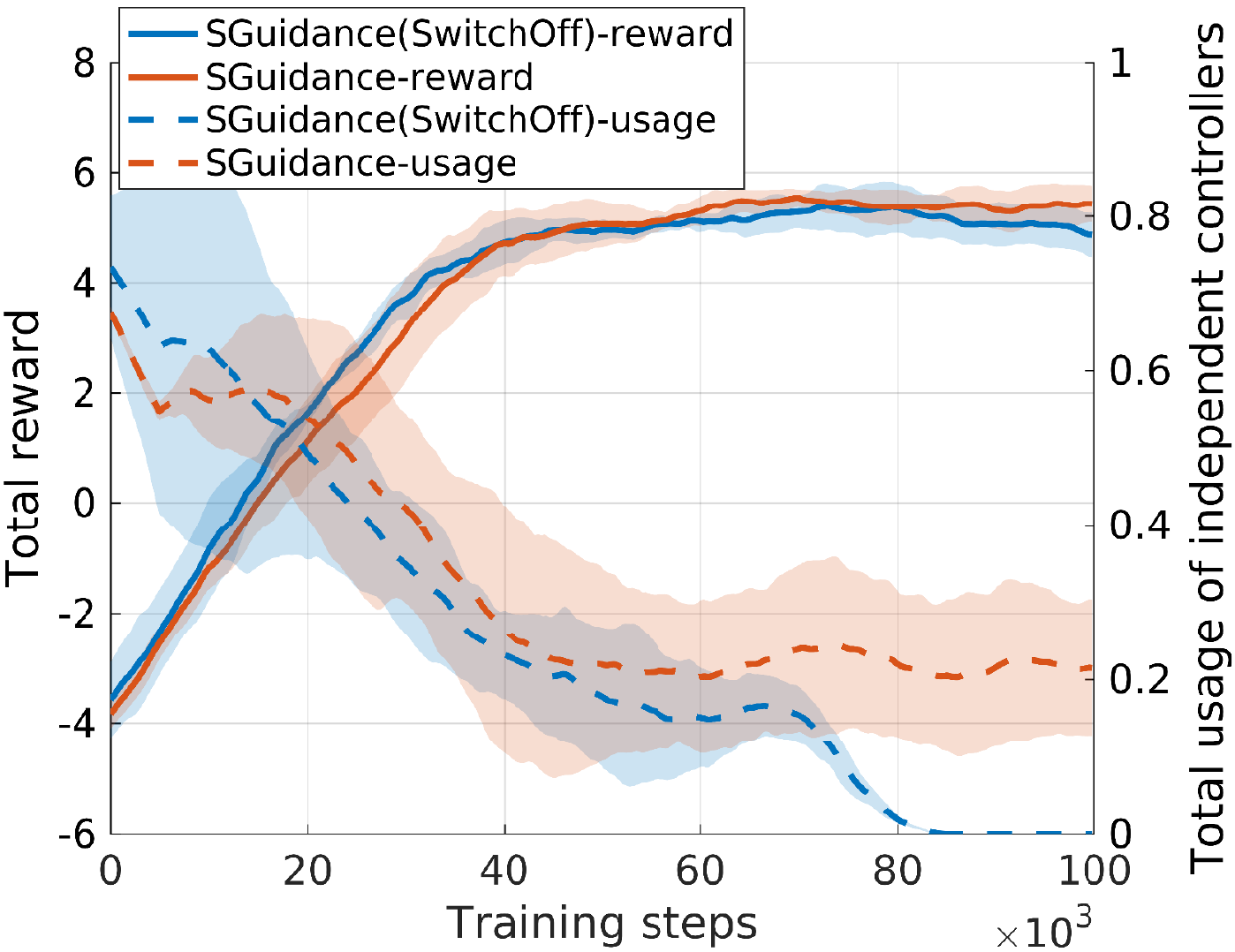}}
    \caption{
    \textbf{(a)} 
    The performance with or without variance reduction techniques. \textbf{DDPG+Switch} is the DDPG model with stochastic switch and control variates(CV). \textbf{DDPG+Switch (no CV)} removes the control variates from \textbf{DDPG+Switch}.
    \textbf{DDPG(Vanilla)} is the conventional DDPG without stochastic switch.
    \textbf{(b)} 
    Turning off independent controllers (PID and OA controllers) after their total usage (dashed lines) falls beneath a threshold (15\%) and the resulting influence on performance.
    }
\vspace{-0.2cm}    
\end{figure}

\paragraph{REINFORCE Variance Reduction}

Fig.\ref{fig:variance reduction} exhibits the performance of DDPG with or without variance reduction techniques.
Due to the fact that REINFORCE algorithm also has the high variance issue, we study the benefits brought by the control variates.
As shown in Fig.\ref{fig:variance reduction}, \textbf{SGuidance (no CV)} has already improved the vanilla DDPG model significantly.
However, by introducing two control variates to reduce the variance of the REINFORCE gradient estimators (Eq.\ref{eq:baseline}), \textbf{SGuidance} is able to further enhance and stabilise the navigation performance, which also illustrates that the high variance issue is certainly influential in the context of deep reinforcement learning.

\paragraph{Turning off Independent Controllers}\label{sec:sensitivity}
This experiment investigates the property of DDPG with stochastic switch that the trained DDPG policies are able to independently carry out navigation with all the heuristic controllers turned off.
In Fig. \ref{fig:switchoff}, the percentage of selecting heuristic controllers by \textbf{SGuidance} are demonstrated with dashed lines. 
Their usage drops quickly and becomes stable after approximately 60k training steps.
This is because that the DDPG controller has already reached a comparable policy as other controllers.
Therefore we turn off both PID and OA controllers when their usage is beneath 15\% to study the performance of isolated DDPG.
Instead of abruptly shutting down the controllers, we monotonically diminish the probability of using a proposed action of heuristic controllers to zero within 10k training steps when they are selected by the switch.
As the result, turning off the heuristic controllers only slightly affect the navigation performance which supports the independent navigation capability of trained DDPG policies.

\subsection{Navigation in Real World Environment}

In this experiment, we quantitatively analyse the performance of our model applied in real world environments.
The model is trained in a simulated world built by \textit{ROS Gazebo} ( Fig.\ref{fig:gazebo}), and directly transferred into real world scenario without any fine tuning in order to verify the effectiveness and strong generalisation of the model.

\begin{figure}[!tb]
\centering
\subfigure[Real world environment]{\label{fig:real_world}\includegraphics[width=40mm]{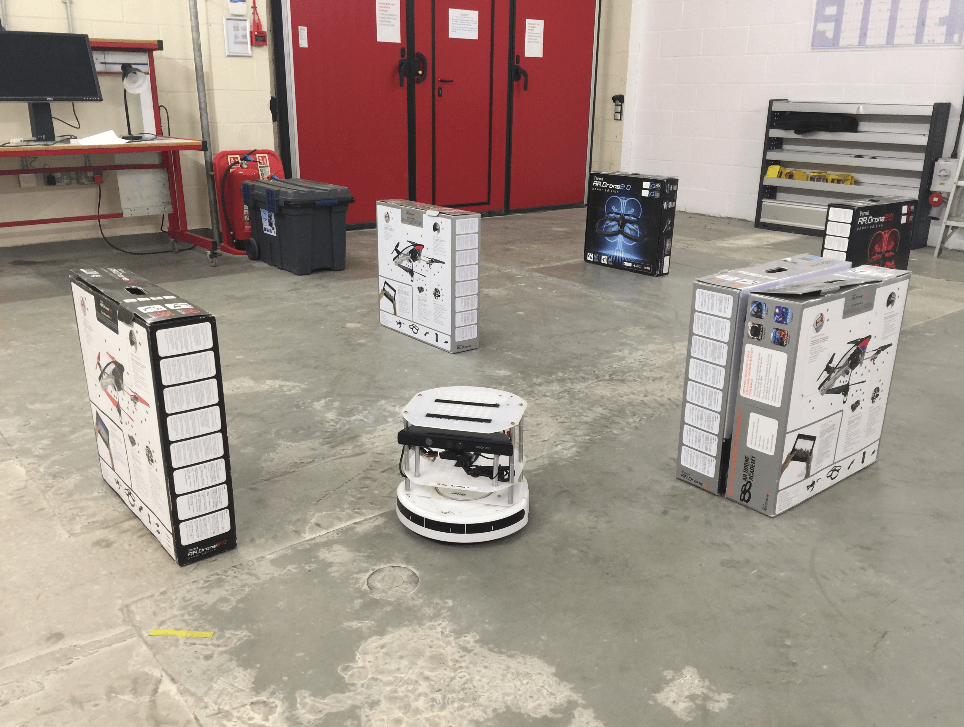}}
\hspace{1.5cm}
\subfigure[Destinations and navigation trajectory]{\label{fig:real_world_traj}\includegraphics[width=45mm]{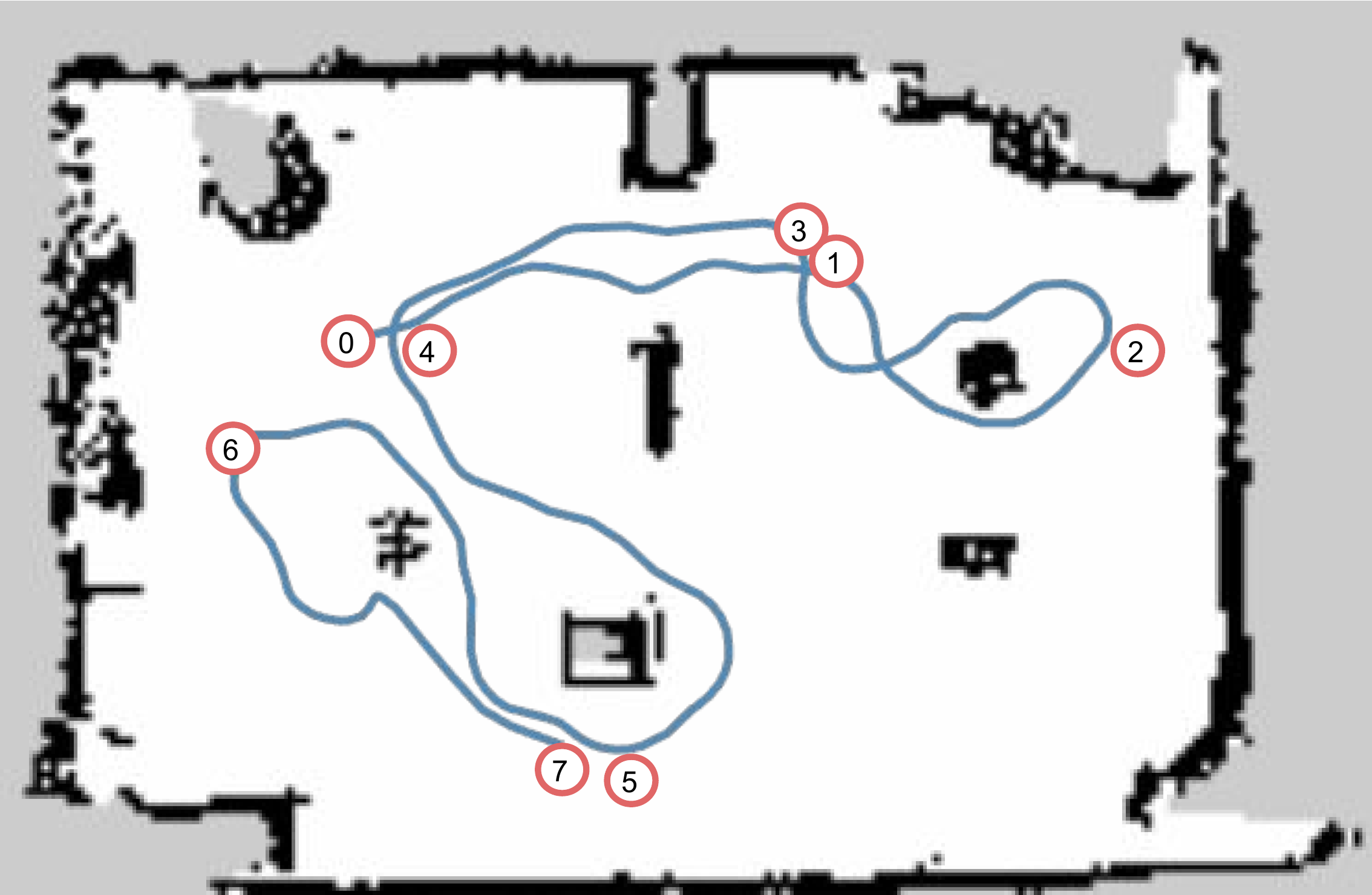}}
\caption{
\textbf{(a)} The real world scenario. A turtlebot is used as the mobile platform and several boxes are placed in the room as obstacles.
\textbf{(a)} The room layout and obstacles are the black areas. The blue curve represents the trajectory of the robot and the goals are plotted with red circles where the number indicates the sequence.
}
\label{fig:real world traj}
\vspace{-0.5cm}
\end{figure}

A Turtlebot 2 robot mounted with a Kinetic depth camera is used as the mobile platform. Unlike the observation from laser scanner which is simulated in \textit{ROS Stage}, the dimension of state space for using a depth camera is dramatically increased. Therefore, a 3-layer convolutional neural network is employed (as in Fig.\ref{fig:Network}) to provide geometric representations.
Other inputs, i.e. velocity and goal location, are concatenated with the geometric representation into a dense input representation. 

Since the ground truth of the robot locations is not available in the real world environment, we apply the off-the-shelf AMCL ROS package\footnote{http://wiki.ros.org/amcl} for providing the estimation of the robot location, and calculating the destination position in the local coordinate frame. 
In order to improve the localisation accuracy, we record the map of the environment with Gmapping ROS package\footnote{http://wiki.ros.org/gmapping}. 
It is worth mentioning that this global map is not used by the navigation component of the model during training or testing.
The obstacles are laid out in the room as illustrated in Fig.~\ref{fig:real_world}.
The target of this experiment is to employ the learned policy and control the robot to reach several destinations successively without any collision.
As shown in Fig.\ref{fig:real_world_traj},
the trajectory of the robot is plotted as the blue curves which indicates that the robot can smoothly avoid all the obstacles and reach each target successfully by only learning in simulation with the proposed stochastic guidance model.

\section{Related Work}

Many works have applied DRL on robotic problems, e.g. navigation [20, 26, 14, 20] and manipulation [6, 25]. 
Since most of the robotics problems involve continuous control, policy based approaches such as policy gradient [19] or actor-critic method, e.g. DDPG [10], are widely used as the conventional approaches. 
Introducing positive bias is a common approach for alleviating the issue. [15] assigns higher weights to the data where the model has less confidence to improve the efficiency of sample usage. 
[7] leverage the concept of information gain when exploring new policies. 
Unlike above approaches where the bias are tightly merged into the models, our framework incorporates extra knowledge as stochastic guidance without imposing any change to the underlying approach. 

Thompson Sampling [1] shares the similar spirit of our framework that the model learns to switch among different controllers.
The difference is that, instead of explicitly calculating the posterior for updating in Thompson Sampling, our framework directly employs neural networks to construct the latent distributions which are trained jointly with the DDPG component by backpropagation. 
The advantage is that the switch function can be easily built and conditioned on all of the sensor inputs so that it chooses different controllers according to different contexts/conditions. 
In addition, the target of our framework is more focused on training a better DDPG component, which is able to benefit from the low-variance gradient estimator due to the better samples generated by the stochastic switch. 

In [9], Leonetti et al. investigated a low level integration of RL and external controllers where the RL algorithm only explores with feasible actions provided by the planner, these heuristics can not be discarded, both for training and testing. Therefore, the performance of the learner very depends on, if not limited by, the capability of the heuristics. 
By contrast, in our framework, DDPG can explore the full action space by itself alongside the guidance during the whole training process and can eventually work independently.

\section{Conclusion}\label{sec:conclusion}

This paper proposes an new framework for effectively incorporating heuristic knowledge to overcome the high variance issue in learning DDPG.
The experiments demonstrate that
the stochastic switch allows an agent to balance the learning from exploration or heuristics, which significantly bootstraps the performance of navigation that surpasses state-of-the-art baseline models.  
More interestingly, the DDPG component remains independent and can be tested in isolation from other controllers. 
By transferring the policies into real world, the robot is able to successfully carry out navigation task, which indicates the robustness and strong generalisation of our proposed framework.

\section*{References}

\small

[1] Shipra Agrawal and Navin Goyal. Analysis of thompson sampling for the multi-armed banditproblem. InConference on Learning Theory, pages 39–1, 2012.

[2] Karl Johan Åström and Tore Hägglund.PID controllers: theory, design, and tuning, volume 2.Instrument society of America Research Triangle Park, NC, 1995.

[3] Yan  Duan,  Marcin  Andrychowicz,  Bradly  Stadie,  Jonathan  Ho,  Jonas  Schneider,  IlyaSutskever, Pieter Abbeel, and Wojciech Zaremba. One-shot imitation learning.arXiv preprintarXiv:1703.07326, 2017.

[4] Dieter Fox, Wolfram Burgard, and Sebastian Thrun. The dynamic window approach to collisionavoidance.IEEE Robotics and Automation Magazine, 4(1):23–33, 1997.

[5] Brian P Gerkey and Kurt Konolige.  Planning and control in unstructured terrain.  InICRAWorkshop on Path Planning on Costmaps, 2008.

[6] Shixiang Gu, Ethan Holly, Timothy Lillicrap, and Sergey Levine. Deep reinforcement learningfor robotic manipulation with asynchronous off-policy updates. InRobotics and Automation(ICRA), 2017 IEEE International Conference on, pages 3389–3396. IEEE, 2017.

[7] Rein Houthooft, Xi Chen, Yan Duan, John Schulman, Filip De Turck, and Pieter Abbeel. Vime:Variational information maximizing exploration. InAdvances in Neural Information ProcessingSystems, pages 1109–1117, 2016.

[8] Mohammad Khan, Shakir Mohamed, Benjamin Marlin, and Kevin Murphy. A stick-breakinglikelihood for categorical data analysis with latent gaussian models. InArtificial Intelligenceand Statistics, pages 610–618, 2012.

[9] Matteo Leonetti, Luca Iocchi, and Peter Stone.  A synthesis of automated planning and rein-forcement learning for efficient, robust decision-making.Artificial Intelligence, 241:103–130,2016.

[10] Timothy P Lillicrap, Jonathan J Hunt, Alexander Pritzel, Nicolas Heess, Tom Erez, YuvalTassa, David Silver, and Daan Wierstra. Continuous control with deep reinforcement learning.International Conference on Learning Representations, 2016.

[11] Yishu Miao, Edward Grefenstette, and Phil Blunsom. Discovering discrete latent topics withneural variational inference. InInternational Conference on Machine Learning, pages 2410–2419, 2017.

[12] Andriy Mnih and Karol Gregor. Neural variational inference and learning in belief networks.arXiv preprint arXiv:1402.0030, 2014.

[13] Volodymyr Mnih, Koray Kavukcuoglu, David Silver, Andrei A Rusu, Joel Veness, Marc GBellemare,  Alex Graves,  Martin Riedmiller,  Andreas K Fidjeland,  Georg Ostrovski,  et al.Human-level control through deep reinforcement learning.Nature, 518(7540):529–533, 2015.10

[14] Fereshteh Sadeghi and Sergey Levine. (cad)2rl: Real single-image flight without a single realimage.Robotics: Science and Systems, 2017.

[15] Tom Schaul, John Quan, Ioannis Antonoglou, and David Silver. Prioritized experience replay.arXiv preprint arXiv:1511.05952, 2015.

[16] Jayaram Sethuraman.  A constructive definition of dirichlet priors.Statistica sinica, pages639–650, 1994.

[17] David Silver, Aja Huang, Chris J Maddison, Arthur Guez, Laurent Sifre, George Van Den Driess-che, Julian Schrittwieser, Ioannis Antonoglou, Veda Panneershelvam, Marc Lanctot, et al. Mas-tering the game of go with deep neural networks and tree search.Nature, 529(7587):484–489,2016.

[18] Reid Simmons and Sven Koenig.  Probabilistic robot navigation in partially observable en-vironments.  InInternational Joint Conference on Artificial Intelligence, volume 95, pages1080–1087, 1995.

[19] Richard S Sutton, David A McAllester, Satinder P Singh, and Yishay Mansour. Policy gradientmethods  for  reinforcement  learning  with  function  approximation.   InAdvances  in  neuralinformation processing systems, pages 1057–1063, 2000.

[20] Lei Tai, Giuseppe Paolo, and Ming Liu. Virtual-to-real deep reinforcement learning: Continuouscontrol of mobile robots for mapless navigation.Intelligent Robots and Systems (IROS), 2017IEEE/RSJ International Conference on, 2017.

[21] Gerald Tesauro. Temporal difference learning and td-gammon.Communications of the ACM,38(3):58–68, 1995.

[22] Christopher JCH Watkins and Peter Dayan.  Q-learning.Machine learning, 8(3-4):279–292,1992.

[23] Ronald J Williams. Simple statistical gradient-following algorithms for connectionist reinforce-ment learning.Machine learning, 8(3-4):229–256, 1992.

[24] Linhai Xie, Sen Wang, Andrew Markham, and Niki Trigoni. Towards monocular vision basedobstacle avoidance through deep reinforcement learning.arXiv preprint arXiv:1706.09829,2017.

[25] Ali Yahya, Adrian Li, Mrinal Kalakrishnan, Yevgen Chebotar, and Sergey Levine. Collectiverobot reinforcement learning with distributed asynchronous guided policy search. InIntelligentRobots and Systems (IROS), 2017 IEEE/RSJ International Conference on, pages 79–86. IEEE,2017.

[26] Yuke Zhu, Roozbeh Mottaghi, Eric Kolve, Joseph J Lim, Abhinav Gupta, Li Fei-Fei, and AliFarhadi. Target-driven visual navigation in indoor scenes using deep reinforcement learning. InRobotics and Automation (ICRA), 2017 IEEE International Conference on, pages 3357–3364.IEEE, 2017

\end{document}